\documentclass[11pt,twocolumn]{article}
\usepackage[]{fullpage}
\usepackage{times}
\usepackage{color}
\usepackage{float}
\usepackage{graphicx}
\usepackage{mathtools}
\usepackage{bbm}
\usepackage{calc}
\usepackage{titlesec}
\usepackage{url}
\usepackage{tikz}
\usepackage{array}
\usepackage{subcaption}
\usepackage{multirow}
\usepackage{booktabs}
\usepackage{url}
\usepackage{breqn}
\usepackage{IEEEtrantools}

\usepackage{breakurl}
\usepackage[breaklinks]{hyperref}

\usetikzlibrary{matrix}
\usetikzlibrary{positioning}

\newcommand{\xhdr}[1]{\vspace{0.2mm}\noindent{{\bf #1.}}}

\newcommand{\ie}{\emph{i.e.}}

\newcommand{\eg}{\emph{e.g.}}

\newcommand{\apname}{{\tt {\small Tetra}}}

\newcolumntype{C}[1]{>{\centering\let\newline\\\arraybackslash\hspace{0pt}}m{#1}}
\newcolumntype{L}[1]{>{\raggedright\let\newline\\\arraybackslash\hspace{0pt}}m{#1}}

\begin{document}

\date{}
\title{The Case for Temporal Transparency: Detecting Policy Change Events in Black-Box Decision Making Systems}

\author{
Miguel Ferreira, Muhammad Bilal Zafar, Krishna P. Gummadi
\\
MPI-SWS, Germany
\\
\{miferrei, mzafar, gummadi\}@mpi-sws.org
\\
\\
}

\maketitle

\begin{abstract}
Bringing transparency to black-box decision making systems (DMS) has been a topic of increasing research interest in recent years. Traditional active and passive approaches to make these systems transparent are often limited by scalability and/or feasibility issues. In this paper, we propose a new notion of black-box DMS transparency, named, temporal transparency, whose goal is to detect if/when the DMS policy changes over time, and is mostly invariant to the drawbacks of traditional approaches. We map our notion of temporal transparency to time series changepoint detection methods, and develop a framework to detect policy changes in real-world DMS's. Experiments on New York Stop-question-and-frisk dataset reveal a number of publicly announced and unannounced policy changes, highlighting the utility of our framework.

\end{abstract}
\section{Introduction} \label{intro}
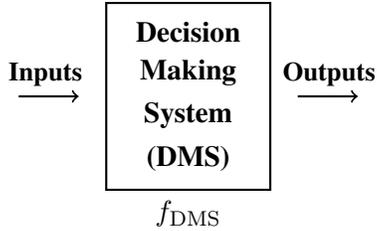
\begin{figure}[t]

\centering
\begin{tikzpicture}
\matrix (A){
\\
};
\matrix (B) [matrix of nodes,row sep=-\pgflinewidth, thick, draw, nodes={rectangle, minimum width=5em},nodes in empty cells, right= 11.5mm of A] {
\textbf{Decision} \\ \textbf{Making} \\ \textbf{System} \\ \textbf{(DMS)} \\
};
\node[anchor=north] at (B.south) {$f_{\mathrm{DMS}}$};
\matrix (C) [matrix of nodes,row sep=-\pgflinewidth, right= 11.5mm of B] {
\\
};

\path[ ->, shorten >= 10pt, thick] (A) edge node[anchor=south]{\small\textbf{Inputs} \quad \quad} (B);
\path[ ->, shorten <= 10pt, thick] (B) edge node[anchor=south]{\quad \small\textbf{Outputs}} (C);
\end{tikzpicture}
\caption{\textbf{The abstraction of a traditional DMS. The decision making policy ($f_{DMS}$) is often unknown. Efforts to bring transparency to DMS focus on inferring $f_{DMS}$ from inputs and outputs.}}
\vspace{-1em}
\label{fig:dms}
\end{figure}

\noindent In modern societies, it is widely accepted that decision
making systems (DMS), particularly those whose outcomes affect
people's lives, need to be {\it transparent}.
However, these decision making systems (example illustration in Figure~\ref{fig:dms}) often act as black-boxes,
where the precise decision making policy or function ($f_{\mathrm{DMS}}$) is not known and hence
the relationship between individual inputs and outputs is not clear.
A number of recent
studies have attempted to bring transparency to black-box
decision making systems, be they driven by machines (e.g., algorithmic
search and recommendation systems~\cite{mislove_price_disc,xray_usenix}) or humans (e.g., stop and
frisk decisions made by police~\cite{goel_frisk,goel_traffic}).
These studies attempt to
reverse-engineer or infer the {\it decision making policy }
(the function $f_{\mathrm{DMS}}$) either by (i) actively auditing the system with
carefully crafted inputs and analyzing the resulting outputs~\cite{mislove_price_disc,xray_usenix}
or by (ii) passively observing the inputs and outputs of the system in
operation~\cite{goel_frisk,goel_traffic}.

The above two broad approaches to bringing transparency have their pros
and cons: (i) active audits can help achieve {\it functional
  transparency}, \ie, learn the behavior of the decision function for
different types of inputs, but they can be expensive and might not
reveal much about the system's behavior under operational conditions
(where inputs are typically drawn from specific probability
distributions over the input space), (ii) passive observations of the
systems' inputs and outputs, on the other hand, can help achieve {\it
  operational transparency}, but they are restricted to analyzing
decision function behavior only on the limited set of operational inputs seen to date.

Against this background, we make the case for a different notion of
transparency that we call {\bf temporal transparency}, where the goal
is to detect {\it when and how the decision making policy
 (the function $f_{\mathrm{DMS}}$) changes over time}. Note that the
objectives of temporal transparency are complementary but different
from those of traditional functional or operational transparency. The
motivating scenarios for temporal transparency are numerous.

\xhdr{1. Monitoring policy change events \& alerting users} Temporal
transparency enables one to track and verify when and how policies of
decision making systems, such as NYPD Stop-question-and-frisk
program (NYPD SQF)
\footnote{\scriptsize{https://en.wikipedia.org/wiki/Stop-and-frisk\_in\_New\_York\_City}}
or Facebook's newsfeed algorithm, have changed over the
years~\cite{facebook_news_feed_2,facebook_news_feed_1}. It would be
possible to monitor whether and when an announced policy change by
public or private organizations has come into effect~\cite{challanges_change_nyu_sqf,chicago_change_sqf}. Furthermore,
any unannounced (or surreptitiously deployed) policy changes can be
detected and used to alert civil liberties and consumer protection
groups to demand greater transparency~\cite{fb_unannounced}.  Later in this paper,
we detect several instances of announced and unannounced policy
changes in NYPD SQF program.

\xhdr{2. Feasible when other transparency approaches aren't} Temporal
transparency can be effective even in scenarios when functional or operational transparency
cannot be achieved. For instance, consider the NYPD SQF program. It is
not feasible to actively audit NYPD's decision making by generating
artificial new inputs (\ie, pedestrians in NYC). One needs to rely on
passively analyzing records of stops maintained by NYPD. But, as NYPD
only records data for pedestrians that have been stopped and does not
record data for all pedestrians that the police are observing, it is
impossibly hard to infer the decision making policy (function) in its
entirety. However, as we show later in the paper, these limited
records are sufficient to achieve temporal transparency, \ie,
robustly detect a variety of policy changes implemented by NYPD over
several years.

\xhdr{3. Finding targets for other transparency approaches} By
detecting the points in time when the decision making policy has
changed, temporal transparency can help focus the more expensive
traditional approaches to transparency (like active audits or passive
input-output analysis) to the short period of time before and after
the policy change events. Focusing transparency efforts on policy
change events can help us better understand the magnitude and effects
of the policy changes on the outcomes of the decision making system.

Intuitively, the basic idea behind detecting changes in decision making policy is as follows: assume we are given a time series of inputs and outputs to the system.
Our task is to detect if/when the decision making policy ($f_{DMS}$) mapping  inputs and outputs, has changed.
Our intuition
for detecting changes in $f_{DMS}$ is to look for temporal
changes in outputs, where inputs remain relatively stationary.

In this paper, we argue that the problem of detecting policy change
events naturally fits existing frameworks for detecting changepoints
in time series.
Time series changepoint detection is a well-studied
problem in statistics, signal processing and machine learning
\cite{barry1992, basseville1993, fearnhead2007,page1954,xiang_bayesian}.
These studies often work with the assumption that any time series with changepoints
consists of observations drawn for different statistical distributions, and at every changepoint,
the distribution that the following observations will be drawn from, changes.
Hence, the changepoint detection problem boils down to recovering the parameters of the underlying
distributions that best explain the observations. As a by-product of this process, one also obtains a list with locations of corresponding changepoints.
However, applying changepoint detection techniques
on real-world datasets, subjected to noise, outliers, seasonal and
weekly patterns, and different magnitudes of the detected changes, is
not a straightforward task.

To tackle these challenges, we developed a
framework called \apname{} (for Temporal Transparency), that builds on  Bayesian changepoint
detection techniques \cite{fearnhead2006,xiang_bayesian}. Specifically,
in order to make the earlier methods robust to transitory disturbances
in the observed features and aiming at detecting only significant policy
shifts, we pose changepoint detection as a \textit{maximum a
  posteriori} (MAP) problem and propose a \textit{dynamic programming}
(DP) solution.
Our framework operates in an unsupervised fashion with the goal of finding the location
of changepoints that best explain the underlying observations. Given an initial set of parameters to tune
the sensitivity of the changepoint detection, it can return a ranked list of changepoints ordered with likelihood
that a certain point indeed corresponds to a policy change. This flexibility can help the system administrator in terms
of adjusting the significance level of the policy changes that are to be detected.

Applying \apname{} on a real-world DMS, NYPD SQF program, provides interesting insights into the policy changes deployed by NYPD.
Specifically, we detect several policy changes deployed by NYPD between 2006 and 2013, including changes announced publicly.

\section{Detecting Policy Change Events} \label{SecFormSet}

In this section, we outline the design of our framework \apname{}, whose goal is to detect policy change events in a DMS.

Let $\mathbf{I}_t$ and $\mathbf{O}_t$ be
 the observable inputs and outputs of a DMS at time $t$.
Let $x_t$ be a statistic computed over $\mathbf{I}_t$ and $\mathbf{O}_t$.
Consider computing  $x_t$ for a period of time $\left[T\right]$.
The set of observed features collected during such time period, $x_1^T = \{x_1, \dots, x_T \}$, can be considered as a time series of data.

The problem at hand consist of finding the optimal set of changes---that is, the number of changes and their respective locations---which best explain the time series $x_1^T$. This setup can leverage time series changepoint detection frameworks. Specifically, we choose to build on  Bayesian probabilistic changepoint detection setups described in \cite{fearnhead2006,xiang_bayesian}
.
Adhering to the notation presented in \cite{fearnhead2006}, the problem above can then be formulated as:

\begin{align} \label{OffProbStat}
	&\text{maximize} \quad P(\tau_1^{m}, m | x_1^T) \\
	&\text{subject to} \quad 1 < \tau_1 < \dots < \tau_m, \nonumber \\
 	& \quad \quad \quad \quad \quad \tau_j - \tau_{j-1} \ge d, \nonumber \\
	& \quad \quad \quad \quad \quad m \in \mathcal{M},  \nonumber
\end{align}
where the optimal parameters ${}^*m$ and ${}^*\tau^{{}^*m} = \{{}^*\tau_1, \dots, {}^*\tau_{{}^*m}\}$  represent the optimal number of changepoints, and their locations, respectively.
$\mathcal{M}$ is defined as a symmetric set around an initial estimation of the number of changes $\hat{m}$ in $x_1^T$.
$\hat{m}$ can be provided by the user as a part of the domain knowledge.
In case the user chooses not to specify it, we consider it to be the result of computing the CUSUM chart (\cite{basseville1993,page1954}) of $x_1^T$ and analyzing its first derivative.

\xhdr{Detecting significant policy regimes}
Notice that, considering we are interested in detecting \textit{significant} policy regimes, we add an additional constraint to the traditional Bayesian changepoint detection frameworks: a minimum length $d$ of each time series segment (policy regime).
The flexibility in choosing $d$ allows for precise tunability of the framework according to a user's definition of \textit{significant} policy regime (which may vary based on the specific application domain being considered).

\xhdr{Solving the MAP Problem}
Given the user-defined likelihood function, $P(x_t^s | m)$,  of the time series data under consideration and $P(\tau_j | \tau_{j+1})$ as the prior distribution of changepoint process, right hand side of Eq. \eqref{OffProbStat} can be decomposed into:

\[
\begin{array}{r@{\;}l}
 &\text{maximize} \quad P(m)P(x_1^T | m) \\
 &\text{subject to} \quad m \in \mathcal{M}
\end{array}
\left(
\begin{array}{r@{\;}l}
\small{\text{maximize}} \quad \small{P(\tau_1^m | x_1^T, m)} \\
\small{\text{subject to}} \quad \small{1 < \tau_1 < \dots < \tau_m} \\
\small{\quad \quad \quad \quad \tau_j - \tau_{j-1} \ge d}
\end{array}
\right).
\]
\begin{equation}\label{OffProbStat2}
\end{equation}

By noticing the sequence of realizations $(\tau_1^m)$ form a discrete-time Markov Chain, the solution to the second term of Eq. \eqref{OffProbStat2} is yielded by a dynamic program, whose recurrence relation for $j \in \left[1:m\right]$ is dictated by:

\begin{align}
T(j, \tau_{j}) \quad = \quad &\text{maximize} \quad P(\tau_{j+1} | \tau_{j}, x_1^T, m) T(j+1, \tau_{j+1}) \nonumber \\
&\text{subject to} \quad \tau_j - \tau_{j-1} \ge d .
\end{align}
In particular, the solution to such dynamic program is given by:
\begin{align}
T(0) \quad = \quad &\text{maximize} \quad P(\tau_1|x_1^T,m)T(1,\tau_1) \quad \nonumber \\
&\text{subject to} \quad \tau_1 \le d.
\end{align}

We fix the prior distribution of number changepoints, $P(m)$, to be a discrete Laplacian distribution (a symmetric distribution) of mean $\hat{m}$ and scale $\beta$.
This choice allows the construction of the set $\mathcal{M}$ to be considered as the range of values around $\hat{m}$ which comprise a percentage $\alpha$ of probability mass function of $P(m)$, and the scale $\beta$ of the distribution translates into the confidence in the initial estimate $\hat{m}$. The tuning parameters ($\alpha, \beta$ and other parameters regarding the analysis of the CUSUM chart) influence the shape of the set $\mathcal{M}$, and therefore the sensitivity of the setting. The joint posterior probability $ P(\tau_{m-j+1} | \tau_{m-j}, x_1^T, m) $ is evaluated as in \cite{fearnhead2006}.

\xhdr{Preprocessing} In order to remove underlying noise in the input time series $x_1^T$, and improve the reliability of the results, we apply the following preprocessing steps to $x_1^T$  before subjecting it to changepoint detection setup outlined above:

\begin{enumerate}

\item \textbf{Outlier Removal.} Outliers are identified through comparison with the shifted moving average, and posteriorly removed. The size of the moving average window, as well as the threshold for outlier identification, must be adapted to each particular problem according to a user's definition of \textit{significant} policy regime, and bearing in mind the variance in the dataset and minimum length $d$ of each time series segment. The removal of outliers helps us ignore the extreme and noisy outputs of the DMS, providing robustness to our setup.
\item \textbf{Feature Scaling.} We scale the time series data $x_1^T$ in a $\left[0-1\right]$ range, in order to simplify the setting of the tuning parameters.
\item \textbf{Filtering and Smoothing} We use a Savitzky-Golay filter \cite{savitzky1964} in order to smooth the input time series data. The parameters of the filter, its window length and the degree of its polynomial fit, are directly related with the sensitivity of the changepoint detection framework.

\end{enumerate}

\section{Detecting Policy Changes in \\ NYPD SQF Program} \label{sec:eval}

In this section, we apply our changepoint detection framework \apname{}
on a dataset related to NYPD SQF program.
The SQF program has been a subject of intense public debate since its conception~\cite{nyc_sqf}, and went through multiple publicly announced policy changes~\cite{nyc_sqf,sqf_decline_2012}.
Our goal in this section is to not only check if/when the policy changes announced by NYPD were implemented but also to explore any unexpected policy changes.

To this end, we model the SQF program as a black-box DMS.
We construct the time series $x_{t}$ from the following observed feature: number of stops made per day under the SQF program.
We assume the time series $x_{t}$ to have been drawn from a Gaussian distribution. Consequently, we model the likelihood function of $x_{t}$ as a Student's t-distribution, whose hyperparameters consist of its maximum likelihood solution (MLE), computed by deploying the expectation maximization (EM) algorithm. We model the prior distribution of changepoint process to be uniform, to reflect the fact that the location of a policy change event is independent of the location of the previous one. Finally we specify the minimum length $d$ of each time series segment to be $15$.

We deployed \apname{} on the stops made during the years 2006 to 2013 (inclusive).
The complete records of the stops made under SQF program are made publicly available at the official website of NYC.
\footnote{\scriptsize{nyc.gov/html/nypd/html/analysis\_and\_planning\\/stop\_question\_and\_frisk\_report.shtml}}
Our framework detected a total of $31$ changepoints. Since the number of detected changepoints is considerably large, we systematically analyzed each of the changepoints. As a result, we were able to separate the changepoints into following categories (listed individually for each year in Table \ref{tab:res}):

\begin{table}[t]
\begin{center}
    \begin{tabular}{ | >{$}C{0.8cm}<{$} | >{$}C{0.95cm}<{$} | >{$}C{0.95cm}<{$} | >{$}C{1.5cm}<{$} | >{$}C{1.0cm}<{$} | >{$}C{1.0cm}<{$} |}
    \hline
    \textbf{Year} & \multicolumn{2}{c|}{\textbf{Seasonal}} & \textbf{Unusual} & \multicolumn{2}{c|}{\textbf{Policy}} \\ \cline{2-3} \cline{5-6}
     & \textbf{S} & \textbf{W} & \textbf{inputs} & \textbf{A} & \textbf{UA} \\
    \midrule
    \hline
     2006 & 1 & 1 & - & - & - \\ \hline
     2007 & 2 & 1 & 1 & - & - \\ \hline
     2008 & 1 & 1 & 1 & - & - \\ \hline
     2009 & - & 2 & - & - & 1 \\ \hline
     2010 & 1 & 1 & - & - & 2 \\ \hline
     2011 & 1 & 1 & 2 & - & 1 \\ \hline
     2012 & 2 & - & 2 & 1 & 1 \\ \hline
     2013 &  -  &  1  &  -  &  -  &  3  \\ \hline
    \end{tabular}
\end{center}
\caption{\textbf{List of detected changepoints from January 01 to 2006, to December 31, 2013. \textbf{S}---Summer; \textbf{W}---Winter; \textbf{A}---Announced, \textbf{UA}---Un-announced.}}
\label{tab:res}
\end{table}

\xhdr{1. Seasonal patterns} These changepoints correspond to slight drops in number of stops made each day around mid-year (summer) and  close to the end of the year (winter). This pattern persists for almost all of the years considered for the analysis. $16$ out of the $31$ detected changepoints fall under this category.

\xhdr{2. Unusual input changes} These changepoints \textit{potentially} correspond to unusual changes in everyday pedestrian population of NYC. For example, we detect a changepoint on October 29, 2012, marking a consistent drop in number of stops made per day until the next changepoint on November 10, 2012. This drop is most probably due to
Hurricane Sandy and its aftermath~\cite{sandy_emergency}.
In fact, on the day when the changepoint occurred---October 29, 2012---the number of stops made over the city is merely $193$,
as compared to an average of $1147$ stops per day for the previous week.
 Similarly, a changepoint marking an increase in number of stops per day on September 22, 2011 could potentially correspond to Occupy Wall Street Movement, that started on September 17~\cite{occupywallst_about}.
In total, $6$ out of $31$ changepoints map to this category.

\xhdr{3. (Un)Announced Policy changes} The changepoints that correspond to neither of the above two categories were likely caused by policy changes implemented by NYPD (because they cannot be explained by input changes).
 For example, we detect a drop in the number of stops made per day starting March 26, 2012. This change is in fact a consequence of a \textit{publicly announced} policy change implemented by NYPD, where,  `increased training' and staffing in `high impact' zones results in an overall decline in number of stops \cite{sqf_decline_2012}. Detection of this changepoint highlights the utility of our framework in verifying the policy changes announced by the governing entities.

\begin{figure}[t]
  \begin{subfigure}{0.55\textwidth}
  \centering
    \includegraphics[height=0.8\columnwidth, angle=-90]{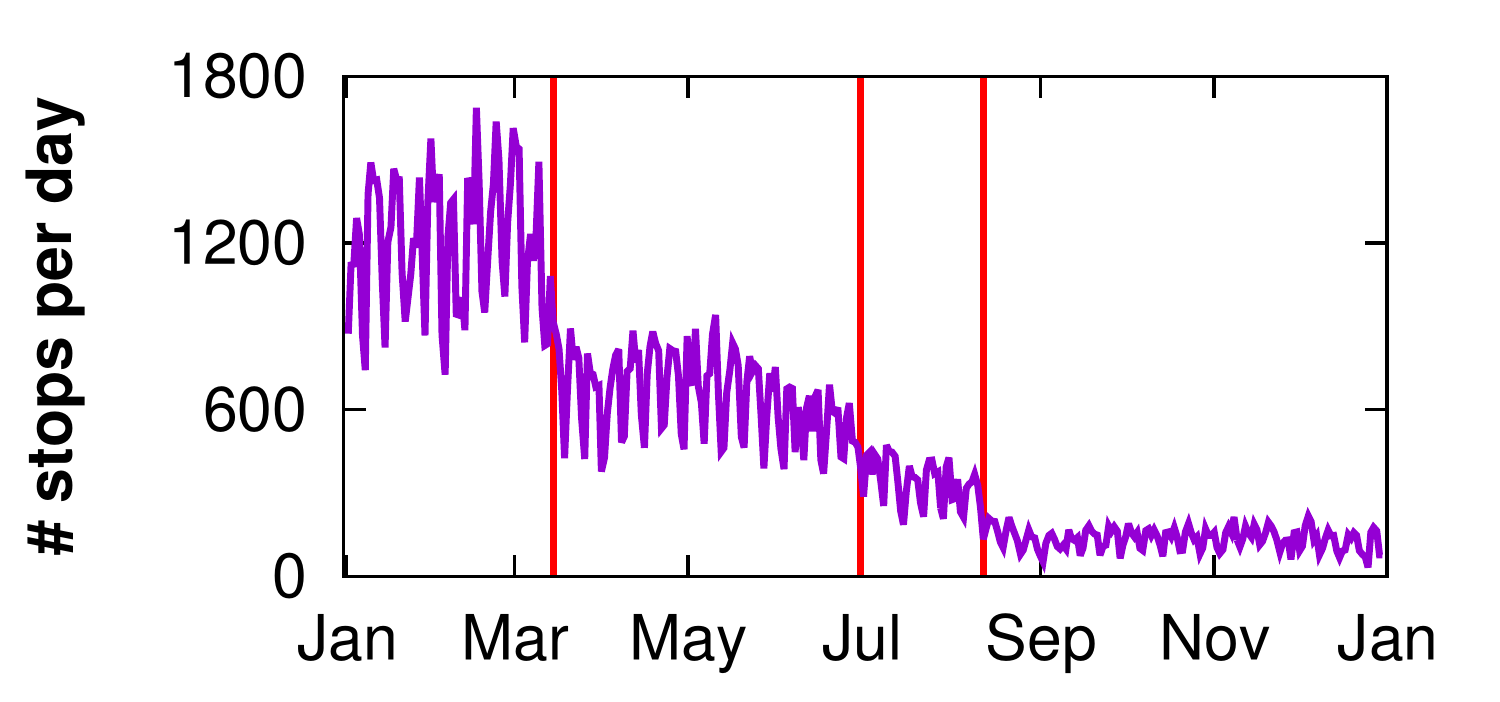}
  \end{subfigure}
  \begin{subfigure}{0.55\textwidth}
  \centering
    \includegraphics[height=0.8\columnwidth, angle=-90]{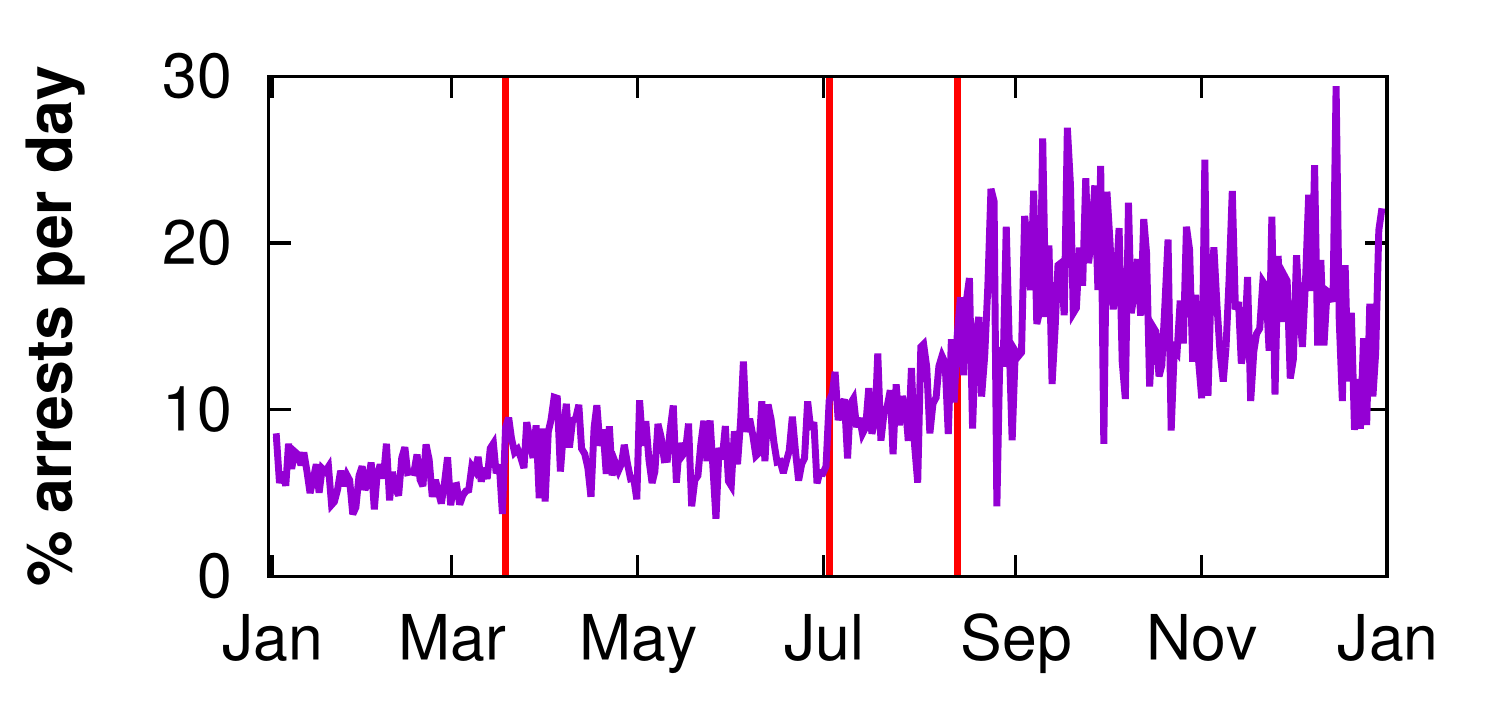}
  \end{subfigure}
\caption{\textbf{Changepoints detected in NYPD SQF data from January 01, 2013 to December 31, 2013.}}
\label{fig:sqf_2013}
\end{figure}

Next, we focus on analyzing changepoints that do not map to a publicly announced policy change.
In particular, we focus the year 2013.
The  changepoint detection framework yields $3$ un-announced changes for this year. Figure \ref{fig:sqf_2013} (top panel) shows the number of stops made per day and the detected changepoints (in the form of vertical lines). Remarkably, this series of changepoints correspond to three \textit{abrupt} policy changes which
 successively brought down the number of stops per day to eventually $10\%$ of the stop rate at the beginning of the year.
 It is important to note that the 2013 SQF program was subject of intense debate during the 2013 Mayoral Election campaign, with  a major candidate denouncing it \cite{blasio_sqf} and a court stating that the SQF policy violated the constitutional rights of the citizens \cite{nyc_judge_2013}.
  Consequently, these variations are likely to be associated with \textit{un-announced} policy adjustments resulting from these events.

In addition to studying the number of stops, we also analyzed the percentage of stops leading to arrests per day in 2013. The changepoint analysis framework detects three changepoints presented in Figure \ref{fig:sqf_2013} (bottom panel), close to the changepoints detected in the stop-rate analysis. This clear mapping between the changepoints yielded by both observed features reveals a systematic change in SQF policy by NYPD, indicating that the policy change did not just concern the number of stops per day, but also the \textit{nature} of the stops.

\section{Conclusion and Future Work} \label{DiscuFutWork}

In this paper we made the case for temporal transparency where the goal is to detect when and how the DMS policy changes over time.
We built a framework \apname{} using prior advances in Bayesian changepoint detection. Applying \apname{}  on a real-world dataset shows
that it can systematically detect possible policy change events in practice.
In the future we hope to generalize our framework to apply it to a broader range of real world DMS's. More specifically, we plan to address the following points:

\noindent
\xhdr{1} The current implementation relies on an `offline' setting that needs access to the whole time series data to detect possible changepoints. Hence, the framework cannot be deployed on streaming datasets, where one might want to detect changepoints on the fly, \eg, Facebook newsfeed algorithm. We are currently expanding it to
incorporate an `online' setting in order to cater to such scenarios.

\noindent
\xhdr{2}
As shown in Section \ref{sec:eval}, analyzing the structure of policy changes by jointly considering multiple observed features (number of stops, percentage of stops leading to arrests, in parallel) can provide more insights into how the DMS interplays with different features, hence revealing more information about policy changes.
To address this point, we plan to generalize our framework to multi-variate feature spaces.

\bibliographystyle{abbrv}

\end{document}